\documentclass[letterpaper, 10 pt, conference]{ieeeconf}

\IEEEoverridecommandlockouts     
\usepackage{hyperref}
\usepackage{pdfpages}
\usepackage{pdflscape}
\usepackage{textcomp}

\usepackage[font=scriptsize,caption=false]{subfig}

\makeatletter
\let\NAT@parse\undefined
\makeatother
\usepackage[numbers]{natbib}

\usepackage{graphicx}
\graphicspath{{figures/}}

\usepackage{graphics} 
\usepackage{epsfig} 
\usepackage{times} 
\usepackage{amsmath} 
\usepackage{amssymb}  
\usepackage{lipsum,adjustbox}
\usepackage{authblk}
\usepackage{booktabs} 
\usepackage{multirow}
\title{\LARGE \bf
Skill Generalization with Verbs
}

\author{Rachel Ma$^{1*}$,
Lyndon Lam$^{2}$, Benjamin A. Spiegel$^{1}$,
Aditya Ganeshan$^{1}$, \\

Roma Patel$^{3}$, Ben Abbatematteo$^{1}$, David Paulius$^{1}$, 
Stefanie Tellex$^{1}$, George Konidaris$^{1}$
\thanks{$^{1}$Brown University, Providence, RI, USA.}
\thanks{$^{2}$California State Polytechnic University, Pomona, CA, USA.}
\thanks{$^{3}$DeepMind, work done at Brown University, Providence, RI, USA}
\thanks{$^{*}$Corresponding Author (\textit{Email}: \texttt{rachelm8@mit.edu})}
\thanks{${^\dagger}$Code, dataset info and demo videos can be found at: \url{https://rachelma80000.github.io/SkillGenVerbs/}.. Also ``© 2023 IEEE.  Personal use of this material is permitted.  Permission from IEEE must be obtained for all other uses, in any current or future media, including reprinting/republishing this material for advertising or promotional purposes, creating new collective works, for resale or redistribution to servers or lists, or reuse of any copyrighted component of this work in other works.”}}

\begin{document}
\maketitle
\thispagestyle{empty}
\pagestyle{empty}

\begin{abstract}
 
It is imperative that robots can understand natural language commands issued by humans. Such commands typically contain verbs that signify what action should be performed on a given object and that are applicable to many objects. We propose a method for generalizing manipulation skills to novel objects using verbs. Our method learns a probabilistic classifier that determines whether a given object trajectory can be described by a specific verb. We show that this classifier accurately generalizes to novel object categories with an average accuracy of 76.69\% across 13 object categories and 14 verbs. We then perform policy search over the object kinematics to find an object trajectory that maximizes classifier prediction for a given verb. Our method allows a robot to generate a trajectory for a novel object based on a verb, which can then be used as input to a motion planner. We show that our model can generate trajectories that are usable for executing five verb commands applied to novel instances of two different object categories on a real robot. 

\end{abstract}

\section{Introduction}

Robots that interact with humans should be equipped with the means to interpret and follow commands in natural language. Manipulation commands are commonly expressed as verbs applied to a given object. 
We therefore propose that robots that can efficiently learn how to perform various manipulation tasks from natural language commands must be able to generate a motor skill that matches a given verb, and apply it  to manipulate a novel object. For instance, opening a door is similar to opening a microwave; therefore, a robot that has learned a skill appropriate for the verb ``open" applicable to a door should be able to: 1) know what an ``open" microwave looks like given a ``closed" microwave, and 2) execute the ``open" action on a microwave with minimal additional learning. However, most works in natural language grounding and generalization either do not apply multiple actions across multiple object categories \citep{hewlettteaching,paulius2020motion,EisnerZhang2022FLOW}, assume robots know goal states for primitive verbs \citep{ahn2022can,sharma2021skill}, or rely on demonstration data \citep{sharma2021skill,eisermann2021generalization,ahn2022can, kollar2014grounding, cui2023no, sugiura2007learning}. 

\begin{figure}[t]
\centering
\includegraphics[width=0.85\columnwidth,clip,trim={0.1cm 0.3cm 0.1cm 0.1cm}]{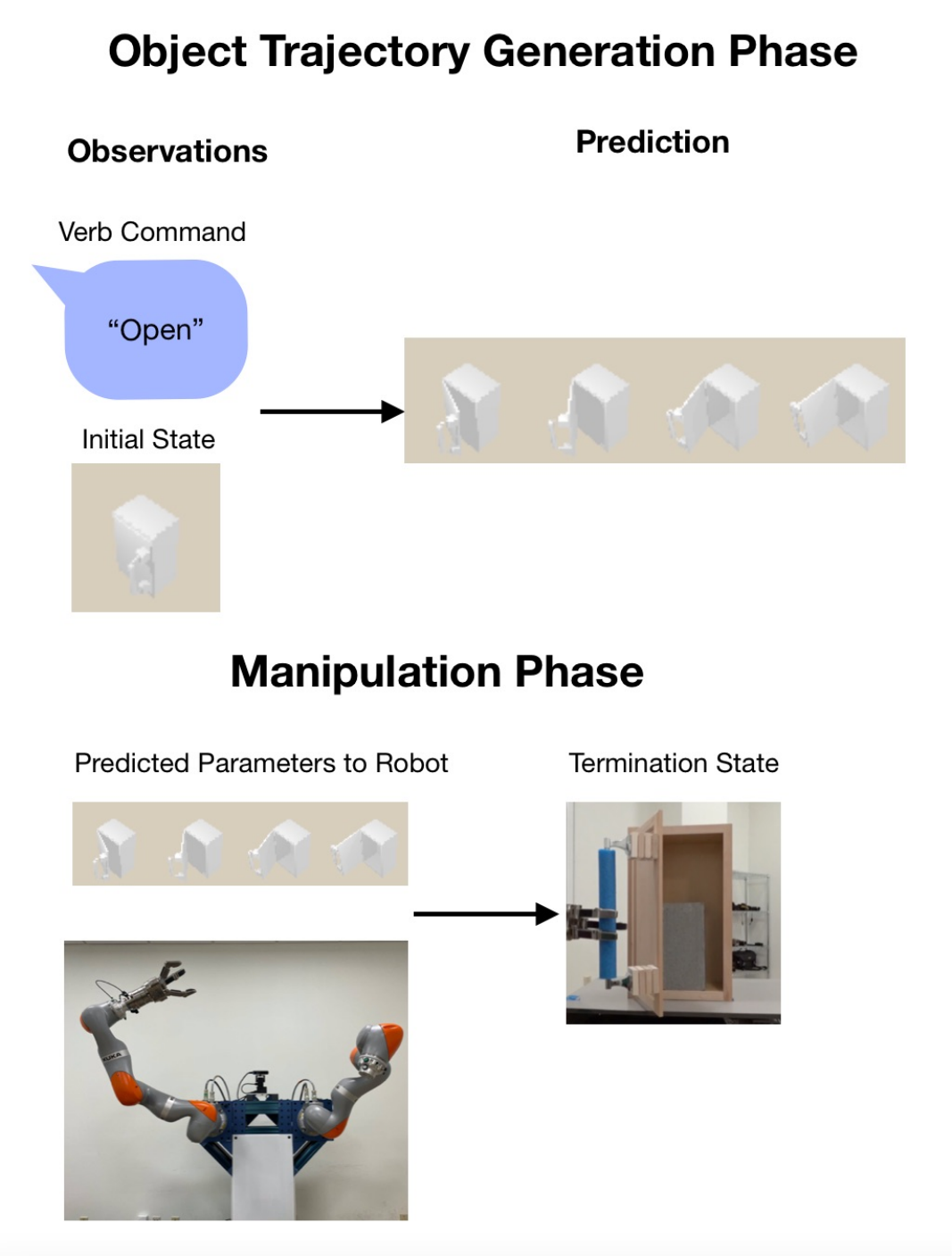}
\caption{Given an observation of an object (before the verb is applied) and a desired verb command, our model generates an object trajectory.}
\label{fig:robotpipeline}
\end{figure}
To address the problem of generalizing verb-labeled skills to novel object categories, we propose a model${^\dagger}$ with two components---a classifier and an optimizer---for producing object trajectories given images of an object to manipulate and a desired verb to execute. The first component is a classifier for the trajectory identification task, where the goal is to identify the correct verb given image snapshots of an object trajectory, allowing us to exploit the effectiveness of neural networks for vision-based tasks. 
This classifier is used by the second component, a policy search algorithm that generates a trajectory from the initial state of an object (before the verb is applied) to a final state (after the verb has been applied), so as to maximize the verb probability estimated by the classifier.

We applied our approach to a list of verbs obtained from VerbNet~\cite{schuler2005verbnet} and objects from the PartNet-Mobility dataset in SAPIEN~\cite{xiang2020sapien}. We observed an average accuracy of 76.69\% when generalizing to an unseen category across 13 object categories and 14 verbs. Along with this classifier, our trajectory optimizer generates plausible object trajectories. To show the efficacy of our model, we show that it can generate trajectories that are usable by a real robot (shown in Figure~\ref{fig:robotpipeline}), through demonstrations with the KUKA LBR iiwa7 robot executing five verb commands applied to novel object instances of two different object categories.

\section{Background and Related Work}
\subsection{Options Framework}
Manipulation tasks are commonly modeled as Markov Decision Processes (MDP)~\cite{Sutton1998}, where given a task, a robot chooses an action and updates its action policy based on reward given its observations of the world. We are particularly motivated by the work of Rodriguez-Sanchez and Patel~\cite{patel2020structure}, which suggests that verbs ground to actions in MDPs. We focus on specifically grounding verbs to motor skills, via the options framework~\cite{sutton1999between}. Each motor skill is modeled by an option $o$ that consists of three components: the option policy $\pi_o$, which is executed and maps low-level states to low-level actions; the initiation set $I_o$, which describe states where the option can be executed; and the termination condition $\beta_o$, which is the probability of the option terminating in each state~\cite{Sutton1998}.

We apply these concepts to develop a model for generating an object trajectory that achieves the intended goal of a verb given visual input of the object. Given the example of applying the verb ``open" to a door, the initiation set would be the current image of the door, which captures the notion of the door in a closed state. The termination set would be an image of the final state of the door, which captures the notion of a door in its open state---specifically, an image of the door angled ajar. The option policy generates a predicted trajectory of the object when the verb is applied.

\subsection{Verbs}
Verbs play a crucial role in natural language commands. \citet{rappaport2010reflections} propose that verbs can be classified as \textit{manner verbs} that specify the manner of carrying out the action, and \textit{result verbs}, that specify the reaching of a resulting state. Result verbs can be classified further into three categories: \textit{change of state verbs}, which specify a change of state of a property of the object the verb is applied to, \textit{inherently directed motion} verbs, which contain movement in relation to an object, and \textit{incremental theme} verbs, which specify a change in volume or area of object. Following \citet{gao2016physical}, and unlike Hovav and Levin, we consider changes of location, volume, and area to also constitute ``change of state".  

To realize the effects of \textit{change in state verbs}, we propose that at least one of the following is required: termination state (the state of the object after the verb is applied), both initiation (the state of the object before the verb is applied) and termination state, object trajectory, and/or robot arm trajectory. Verbs like open and close can be minimally differentiated by termination state. Verbs like rotate can be minimally differentiated by initiation and termination state by looking at the difference of angles between the two steps, while verbs like throw and toss can be minimally differentiated by the combination of object and robot arm trajectories. 
We focus on \textit{change in state} verbs that can be realized through initiation and termination, and/or object trajectory, where the agent manipulates a physical object.

\subsection{Related Work}

Existing work has focused on grounding language in a visual representation of objects in the world and generalizing manipulation skills for robots. We situate our work between these bodies of work and outline them below.

\textbf{Grounding Language to Vision and Manipulation}:
Many works assume that robots already know the goal state of a given object after applying primitive/atomic verbs. \citet{ahn2022can} and \citet{sharma2021skill} focused on breaking down complex natural language commands into simpler primitive actions or tasks. \citet{ramesh2022hierarchical} created a model for text to image generation, which could be useful for predicting images of objects after manipulation. However, given the text prompt ``closed oven", such a model produces mostly open ovens, and we have found that it has trouble with differentiating between physical states of an object. \citet{paulius2020motion} proposed a motion taxonomy for describing action verbs as binary strings known as motion codes, which can be used to describe action and discern between the meaning of actions in a manipulation-centric embedding space. However, they did not account for manipulation on objects across different categories. 
 
Other work in natural language and robotics  addressed language-conditioned imitation learning \citep{lynch2021language} or trajectory modification with natural language commands \citep{spiegel2021guided,bucker2022robot,cui2023no}. However, these works do not address generalization of tasks or skills across objects or of skills. Rather, they focus on imitation learning, which requires large amounts of demonstration data, 
or active parsing and interpretation of commands from humans during a task to accomplish the skill.

\textbf{Learning Generalizable Skills}: Contrary to our work, where we focus on generalizing skills by the effect or action of the verb, a common approach is generalizing through object articulation. Eisner and Zhang~\cite{EisnerZhang2022FLOW} proposed a model to learn and predict 3D articulation flow for various objects; this output was then used to execute a motion planner to achieve the maximum articulation. However, there is no explicit integration of language nor mention of multiple verbs or skills being applied to each object instance. \citet{abbatematteo2019learning} investigated how to estimate the kinematic model and configuration of novel objects for manipulation; however, they do not explore generalization of skills. \citet{hewlettteaching}, \citet{jang2022bc}, and \citet{sugiura2007learning} incorporated the presence of another human in the environment/loop or human demonstration data, which is inefficient and thus limits the amount of verbs that the approach can handle.
Furthermore, while \citet{hewlettteaching} required a human in the loop of identifying the verb for a demonstrated trajectory, our approach exploits deep neural networks for verb prediction.

 \begin{figure*}[t]
 \vspace{5mm}
 \centering\includegraphics[width=0.75\textwidth, height=6.5cm,clip,trim={0 0.70cm 0 0}]{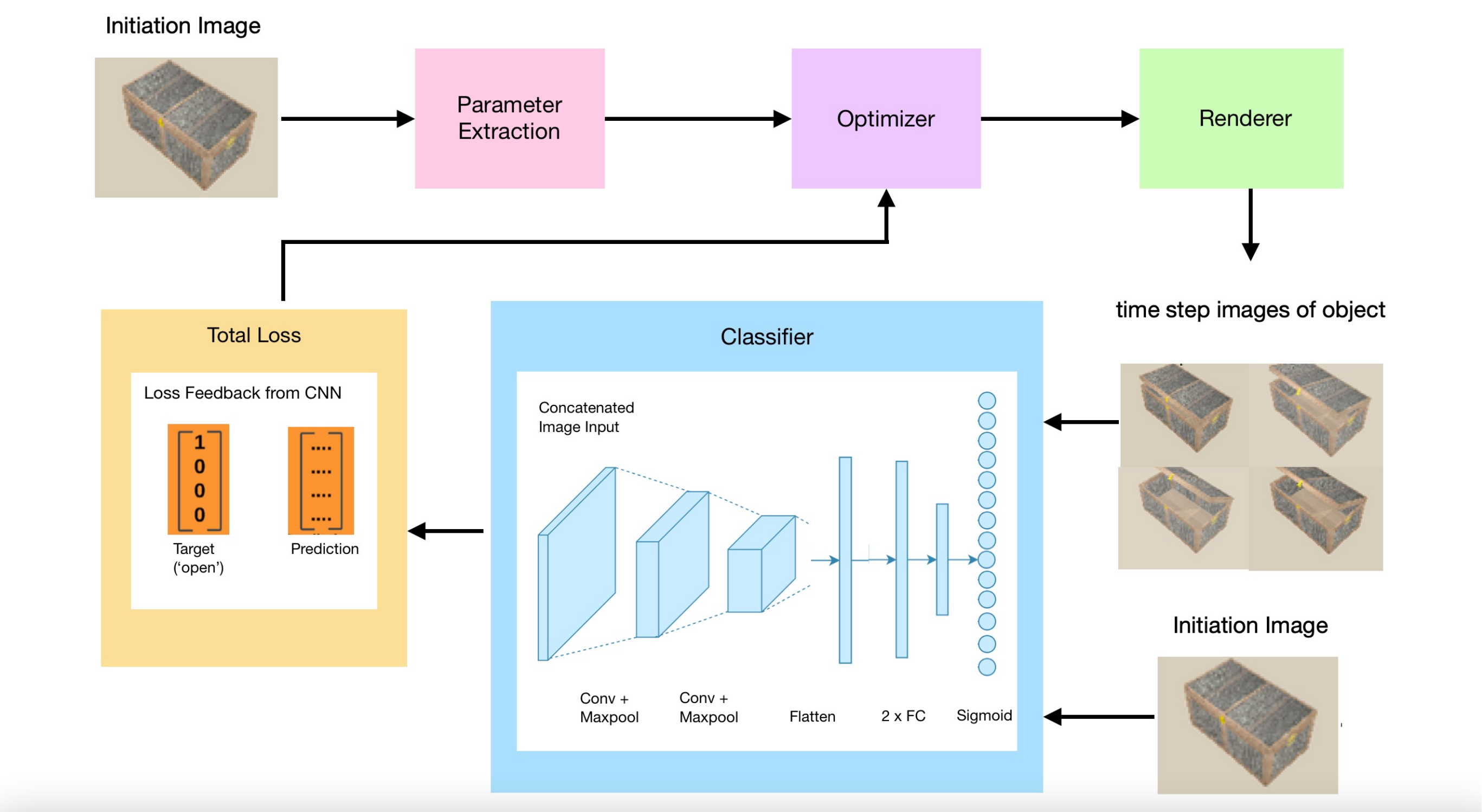}

\caption{Diagram of the planning portion of the proposed model. Parameters of the initiation image are extracted and then manipulated by the optimizer, which relies on the categorical cross-entropy loss calculated on the probabilities from the classifier and the target array. Trajectory timestep images (excluding the initiation image) are rendered and given to the classifier, along with the initiation image for computing the loss.}
\label{fig:model}
\end{figure*}

\section{Skill Generalization With Verbs}
We propose a model that generalizes verb-labeled skills to novel object categories. Our goal is to train a model that takes as input a verb, paired with a kinematic model of an object and its initial state, and outputs a trajectory that can be applied on the object to undergo the effect of that verb. 

In order to realize these \textit{change of state} verbs, robots must be able to: 1) predict the goal/target state of the object if a verb were to be applied, and 2) adapt verb skills (specifically the object trajectory) across multiple (and potentially novel) object categories with minimal additional learning. To achieve these, the model should be able to realize and differentiate the verb that is depicted for a given trajectory, and be able to change the state of the object to achieve the desired verb. We do this through two main components after extracting kinematic parameters for an object instance: a \textit{classifier} that will output a predicted probability for a given verb command and images of the object trajectory, coupled with an \textit{optimizer} which uses this classifier to generate a trajectory from a given initiation state of a novel object to achieve the verb command.  Figure~\ref{fig:model} shows an overview of the planning portion of our model pipeline. 

\subsection{Classifier}
\subsubsection{Architecture}Our experiments use a simple CNN with the following structure: two CNN layers (32 and 64 units respectively with $3 \times 3$ kernel sizes and max pooling layers) and three fully connected layers (64, 32, and $N$ units, where $N$ is the total number of verbs). The inputs to the classifier are sequences of RGB images of a trajectory sequence (where each image has dimensions of $128 \times 128 \times 3$).  Each image of a sequence is concatenated together in the network. The final layer is a softmax layer of $N$ units that will output probabilities of the sequence achieving the effects of $N$ verbs; 

the predicted verb is that with the highest probability value. Other state-of-the-art activity and object recognition methods can be used (refer to Section \ref{sec: Discussion}).

\subsubsection{Training} 
Given $k$ object categories, we perform training and testing in an all-but-one procedure (i.e., $k$-fold cross validation), where we train the classifier using $k-1$ object categories and test the classifier on a unseen $k$-th object category. During training, 80 percent of images from the selected object categories will be used as the train set, and 20 percent will be used as the validation set after shuffling. We use the Adam Optimizer for training.

\subsubsection{Prediction} 
When given an input RGB trajectory sequence for an object instance of an novel object category, the classifier outputs a per-verb probability array.

\subsubsection{Accuracy} 
The accuracy of the classifier is measured by generating predictions on RGB trajectory sequences from the novel test object categories, and comparing the verb label of the most likely prediction with the ground truth verb label associated with the trajectory. 

\subsection{Trajectory Optimizer} 
\label{sec:trajopt}
The second step of our model generates an object trajectory for the desired verb command given a URDF description of the object representing its links and joints. The optimizer searches over trajectories of the degrees of freedom of the object (i.e., 6-DoF pose and articulated state) to maximize the classifier's returned probability for the desired verb. 
We used the Covariance Matrix Adaptation Evolution Strategy (CMA-ES) optimizer algorithm \cite{hansen2016cma}, an efficient state-of-the-art optimization algorithm, to search over object trajectories. 

\subsubsection{Trajectory Parameterization} 
A candidate trajectory is parameterized by a vector of length $(6+n)$, corresponding to the per-timestep change in 6-DoF position and orientation of the object as well as the state of its $n$ articulated joints.
Articulated states are clipped to lie within the joint limits provided in the URDF description. This trajectory is then rendered and scored by the classifier. 

\subsubsection{Measuring Loss} 
The loss minimized by CMA-ES is calculated as the categorical cross-entropy between the verb probability array produced from our CNN classifier and a one-hot target array indicating the target verb.

\subsection{Verb Selection and Data Collection}
To train and evaluate our model, we require visual data of objects undergoing manipulations corresponding to verbs, and their underlying kinematic and geometric descriptions. 

\subsubsection{Visual Data Collection}
We require an environment that provides the following: multiple object categories and multiple instances of objects in those categories, the capability to manipulate parts or the entire object for a variety of verbs, and a way to extract multiple images of the scene. 

We considered datasets such as ALFRED~\cite{ALFRED20}, AI2THOR \cite{kolve2017ai2}, New Brown Corpus \cite{ebert2020visuospatial}, SAPIEN~\cite{xiang2020sapien}, and RL Bench~\cite{james2019rlbench}. We chose the PartNet-Mobility Dataset from SAPIEN due to the presence of 2347 object instances over 46 object categories as URDF files. Each file describes the kinematics (links and joints) and geometry of an object. Our approach performs verb-based manipulation of each object by simulating them in SAPIEN and taking RGB snapshots. For each object-verb pairing, we generate 21 RGB images/snapshots (each 128 by 128) representing 21 timesteps along the object trajectory when the verb is applied to the object. We generate these for our chosen verbs applied for a total of 812 object instances present in the 13 chosen object categories: Box, Dishwasher, Door, Laptop, Microwave, Oven, Refrigerator, Safe, Stapler, Storage Furniture, Toilet, Trash Can, and Washing Machine.  A total of 41688 trajectories are generated. 

We assume maximally distinct initiation and termination states for collecting data needed to train the CNN classifier. For instance, the initiation state for the ``open" verb on door is when the door is completely closed, while the termination state for the ``open" verb on a door is when the door is open to the upper joint limit. We also assume that for ``open" and part-based translation verbs applied to objects with multiple non-fixed joints, only one joint is manipulated. To prevent the model from generating object trajectories where multiple verbs occur simultaneously, data for a ``none" category is generated, which features objects undergoing multiple verbs at once in a single trajectory.

\subsubsection{Verb Selection}
Our requirements when selecting verbs are whether they can be achieved via the URDF files of object instances and if they can be applied to multiple object categories that are present in SAPIEN. With these objectives in mind, we  examined the verbs in VerbNet for \textit{change in state} verbs. 
The final selected verbs are: translation verbs (specifically \textit{Push}, \textit{Pull}, \textit{Raise}, \textit{Lower}, \textit{TranslateLeft} and \textit{TranslateRight}, and \textit{RemoveWhole}---when removing an object from the scene), part-based translation (\textit{RemovePart}---when a single part of the object is removed---and \textit{InsertPart}---when a single part of the object is inserted),
\textit{Open}, \textit{Close},
and rotation verbs (\textit{Roll}, \textit{Turn}, \textit{Flip} by 270 degrees).

\begin{table*}[t]
\vspace{5mm}
\caption{Accuracy by Verb across Unseen Object Categories (trained by $k$-fold cross validation)}
    \centering
        \small
    \begin{tabular}{ l c c c c c c }
    \toprule[1pt]
    \textbf{Object Types} & \multicolumn{6}{c}{\textbf{Verb Accuracy (\%)}} \\
    \cmidrule{2-7}
    & 	\textit{Translate Verbs} & \textit{Open/Close} &	\textit{Remove/Insert Part} &	\textit{Remove Whole}	&	\textit{Rotate Verbs}  &	\textit{None}  \\
    \midrule
    Box	&  $	96.2	\pm	0.7	$ & $	78.3	\pm	7.4	$ & $	75.8	\pm	7.3	$ & $	99.4	\pm	1.5 $ & $	62.3	\pm	5.2 $ & $	85.3	\pm	2.5$\\
    Dishwasher &  $	97.6	\pm	1.5	$ & $	82.8 \pm	5.8	$ & $	98.2	\pm	2.6	$ & $	99.0 \pm	1.1 $ & $	89.4	\pm	1.8 $ & $	91.3	\pm	1.3$\\
    Door&  $	83.2	\pm	3.6	$ & $	37.0	\pm	7.8	$ & $	75.2	\pm	8.6 $ & $	93.1	\pm	5.2 $ & $	47.0	\pm	9.0 $ & $	74.2	\pm	6.4$\\
    Laptop&  $	86.8	\pm	4.4	$ & $	26.4	\pm	11.8	$ & $	67.0	\pm	14.8	 $ & $	92.4	\pm	2.1 $ & $ 27.1	\pm	4.8  $ & $	84.0	\pm	8.7$\\
    Microwave&  $	93.1	\pm	2.3	$ & $	96.4\pm	3.7	$ & $	98.2 \pm	3.0	$ & $	99.0	\pm	2.6 $ & $	80.7	\pm	5.5 $ & $	79.4	\pm	3.0$			 \\
    Oven&  $	96.4	\pm	2.4	$ & $	98.0	\pm	0.6	$ & $	99.0	\pm	1.3	$ & $	100.0	\pm	0.0 $ & $	86.0	\pm	2.3 $ & $	96.4	\pm	1.4$\\
    Refrigerator&  $	89.6	\pm	3.0	$ & $	72.6	\pm	8.8	$ & $	97.1	\pm	1.2	$ & $	96.6	\pm	3.7 $ & $	84.1	\pm	3.2 $ & $	82.6	\pm	4.4$ \\
    Safe&  $	98.8	\pm	1.1	$ & $	89.0	\pm	16.8	$ & $	96.4	\pm	1.1 $ & $	99.4	\pm	1.4 $ & $	84.6	\pm	3.2 $ & $	93.6	\pm	2.9$\\
    Stapler&  $	54.6	\pm	6.1	$ & $	72.6	\pm	8.9	$ & $	63.1	\pm	14.1 $ & $	88.4	\pm	7.6 $ & $	16.1	\pm	5.1 $ & $	47.6	\pm	11.9 $\\
    StorageFurniture&  $	94.1	\pm	3.3	$ & $	61.6	\pm	5.9	$ & $	89.5	\pm 6.0	$ & $	98.9\pm	0.4 $ & $	79.3	\pm	4.5 $ & $	79.9	\pm	10.5 $\\
    Toilet&  $	50.0	\pm	14.7	$ & $	42.4	\pm	15.6	$ & $	57.8	\pm	11.3	$ & $	60.9	\pm	27.5 $ & $	32.2	\pm	7.5 $ & $	55.2	\pm	12.9 $\\
    TrashCan&  $	93.7	\pm	5.5	$ & $	73.5	\pm	11.0	$ &$	81.7	\pm	9.8	$ & $	97.1	\pm	2.0 $ & $	80.9	\pm	2.8 $ & $	90.9	\pm	2.8 $\\
    WashingMachine&  $	87.8	\pm	2.8	$ & $	91.1	\pm	4.2	$ & $	89.2\pm	5.9	$ & $	94.1	\pm	5.3 $ & $	84.6	\pm	4.1 $ & $	63.2	\pm	19.4 $\\
    \bottomrule[1pt]	
    \end{tabular}

    \label{table:AccVerbacrossObCat}
    
    \end{table*}

\section{Experiments}

The aim of our evaluation is to test the hypothesis that our model can successfully transfer verbs to novel object categories, both in simulation and with a real robot.
We do so by selecting object categories for training the classifier, and then selecting a novel object category for evaluation by the classifier and for manipulation with the optimizer.

\subsection{Classifier}
We measure the accuracy of the classifier on RGB image trajectories from the test object categories. The object categories that we use for our experiments are: Box, Dishwasher, Door, Laptop, Microwave, Oven, Refrigerator, Safe, Stapler, Storage Furniture, Toilet, Trash Can, and Washing Machine.

\begin{figure}[!h]
\centering
\includegraphics[width=0.8\columnwidth, height=6cm]{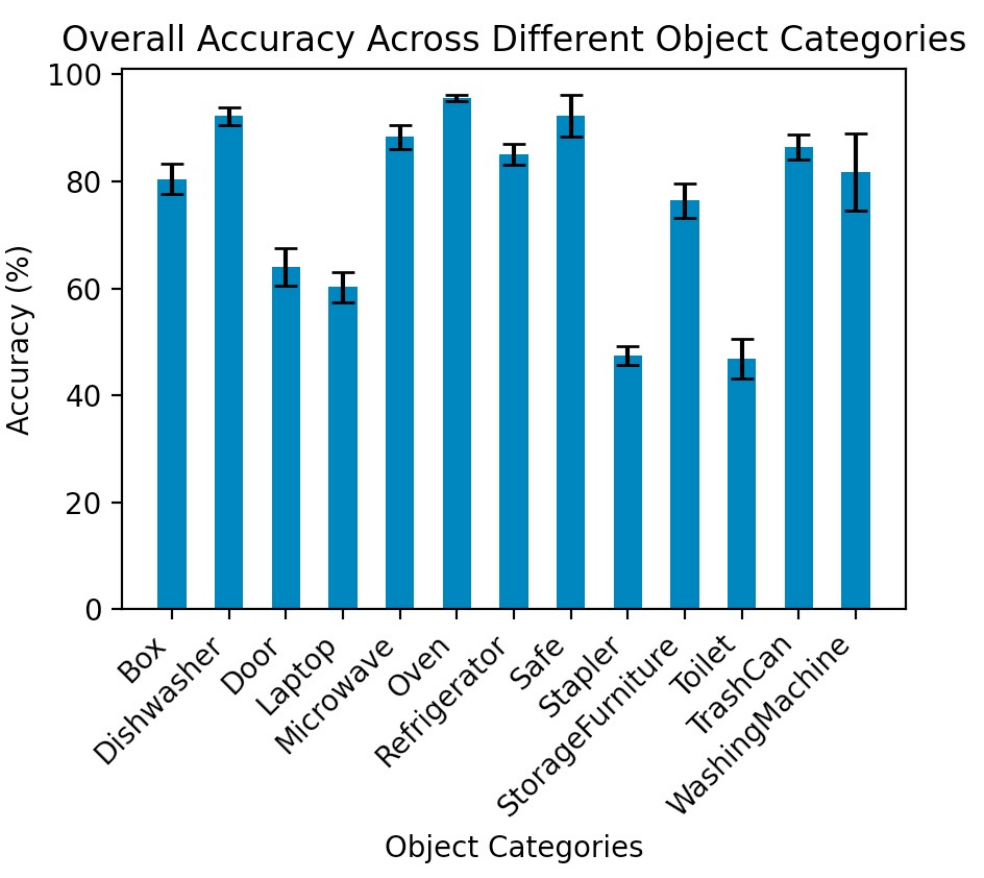}

\caption{Overall Verb Accuracy Across Object Categories. We perform $k-$fold cross validation, where there will be $k-1$ object categories used for training the classifier, and a unseen $k$-th object category reserved for testing. Each of the 13 categories take their turn being the $k$-th category. ``None" verb trajectories are included in training and testing.}
\label{fig:AccOverallVerbacrossObCat}
\end{figure}

\begin{figure}[!h]
\centering
\includegraphics[width=0.9\columnwidth]{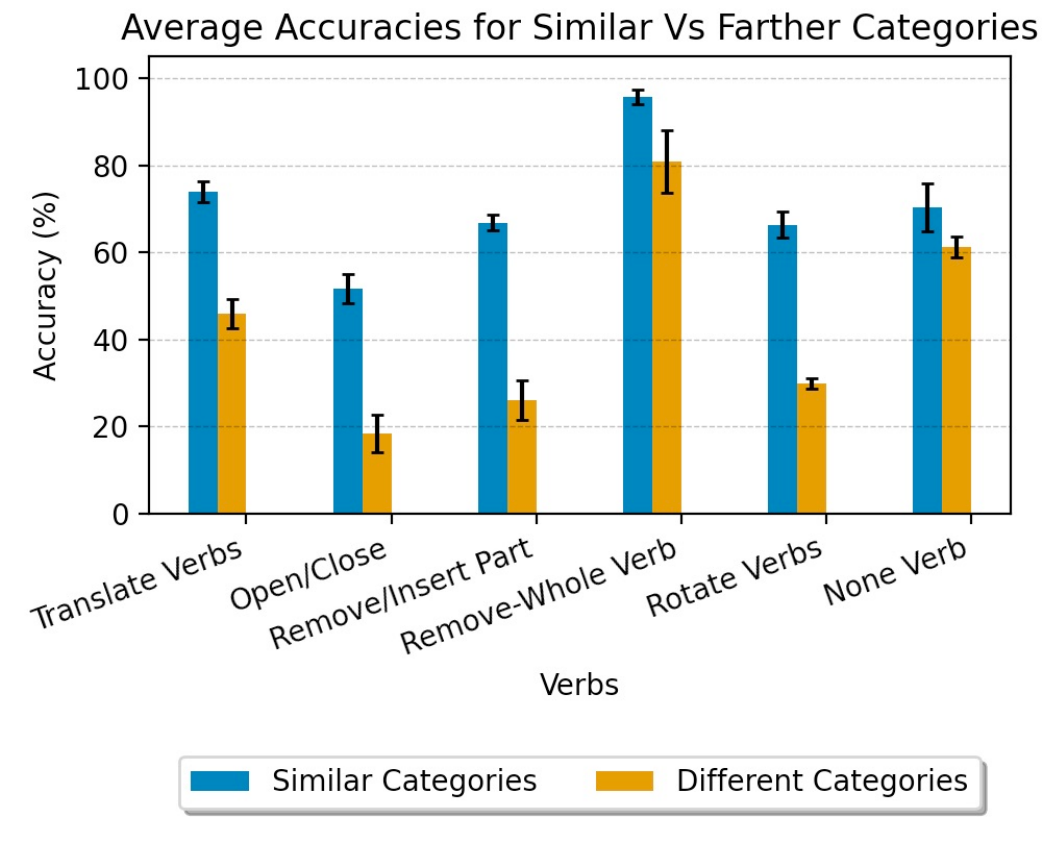}

\caption{Average Accuracies for Similar versus Farther Categories. ``None" verb trajectories are included in training and testing. For testing the ``similar" categories: average of $k$-fold cross validation across categories of Dishwasher, Door, Microwave, Refrigerator, Safe, Washing Machine, where the $k$-th category for testing was one of those, and $k-1$ were training categories. The ``farther" categories: average of $k$-fold cross validation when the $k$-th category for testing is chosen out of Box, Laptop, Stapler, Toilet, and Trash Can, and the training categories were Dishwasher, Door, Microwave, Refrigerator, Safe, Washing Machine.}
\label{fig:Comparison}
\end{figure}

The snapshots selected for use are at regular intervals. We noticed a general increase in accuracy as the number of steps used for the trajectory is increased. We use 5 timesteps as a default for the remainder of our experiments, as we empirically observed that using 5 timesteps provides good accuracy while keeping overall training time relatively low.

We perform $k$-fold cross validation, where there will be $k-1$ object categories used for training the classifier, and an unseen $k$-th object category that will be reserved for testing. The number of epochs is kept constant at 40, and the number of steps is kept at 5 out of the 20 total steps present for each trajectory in the dataset (initiation, Step 5, Step 10, Step 15, and termination snapshots). Overall, the average is 76.69\% accuracy in identifying 14 verbs across 13 object categories. In Figure \ref{fig:AccOverallVerbacrossObCat}, we see the object categories that do the poorest overall are Stapler and Toilet, likely due to their larger differences in shape and usage in comparison to the other object categories. Table \ref{table:AccVerbacrossObCat} shows the accuracy by verb for each object category. In the table, translate verbs include \textit{Lower, Raise, Push, Pull, TranslateLeft,} and \textit{TranslateRight}, while rotate verbs include \textit{Roll, Turn,} and \textit{Flip}. \textit{RemoveWhole} does consistently does well for classification, likely due to a mixture of being a large difference between the initiation and termination states and that the termination state, an object-free environment, appears the same for each object category. In these experiments, the \textit{None} verb trajectories are included in training and testing. 

We conducted another experiment that compares the performance of the classifier when the test object category is similar to the object categories that are used for training, in comparison to a test category that is different. During training, the number of epochs is set to 40 and the number of timesteps is set to 5 (the initiation, Step 5, Step 10, Step 15, and the termination snapshots). \textit{None} verb trajectories are included in training and testing. As seen in Figure \ref{fig:Comparison}, the average accuracy with the $k$-fold cross validation across the object categories Dishwasher, Door, Microwave, Refrigerator, Safe, and Washing Machine, when the $k$-th category is one of those categories, is higher then the $k$-th test category being one of the categories Box, Laptop, Stapler, Toilet, and Trash Can. We believe this is due to the objects sharing similar features of being rectangular and have doors that open in the same orientation. 

\begin{figure}[!h]
\centering
\subfloat[Trajectory Optimizer Result for \textit{Open} Applied to a Novel Cabinet. ]{
\includegraphics[width=0.95\columnwidth]{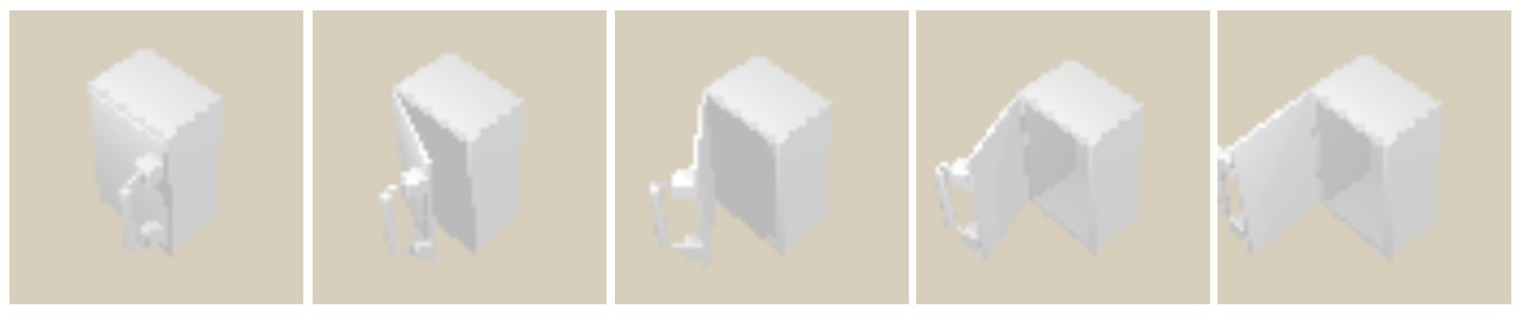}\label{fig:open_cabinet_sim}}

\subfloat[Trajectory Optimizer Result for \textit{TranslateRight} Applied to a Novel Box.]{
\includegraphics[width=0.95\columnwidth]{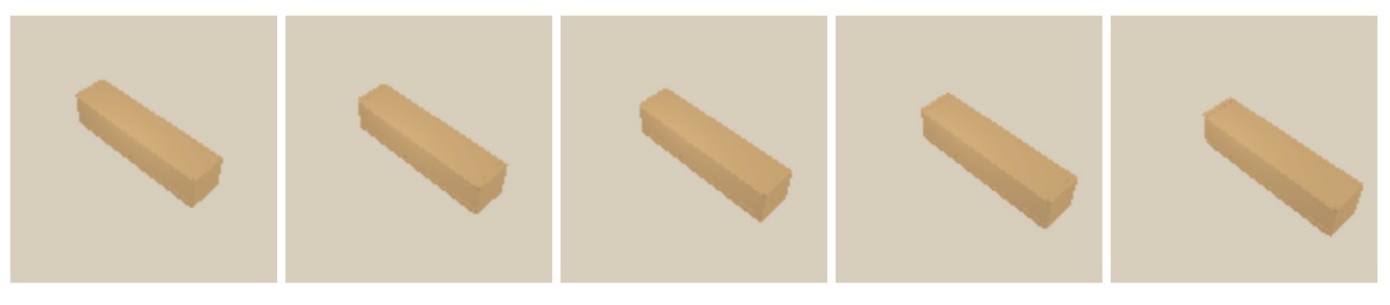}\label{fig:right_box_sim}}

\caption{Examples of trajectory optimizer results. (a) The trajectory optimizer correctly decided to manipulate the joint limit parameter by 0.40 radians for each timestep, thus producing a correct \textit{Open}. (b) The trajectory optimizer correctly decided to manipulate the $y$ parameter by $-0.10$ for each timestep, thus producing a correct \textit{TranslateRight}.}

\label{fig:sim-demo}
\end{figure}
\begin{figure*}[!h]
\vspace{5mm}
\centering
\subfloat[\textit{Open} Applied to a Novel Cabinet]{
\includegraphics[width=1.6\columnwidth]{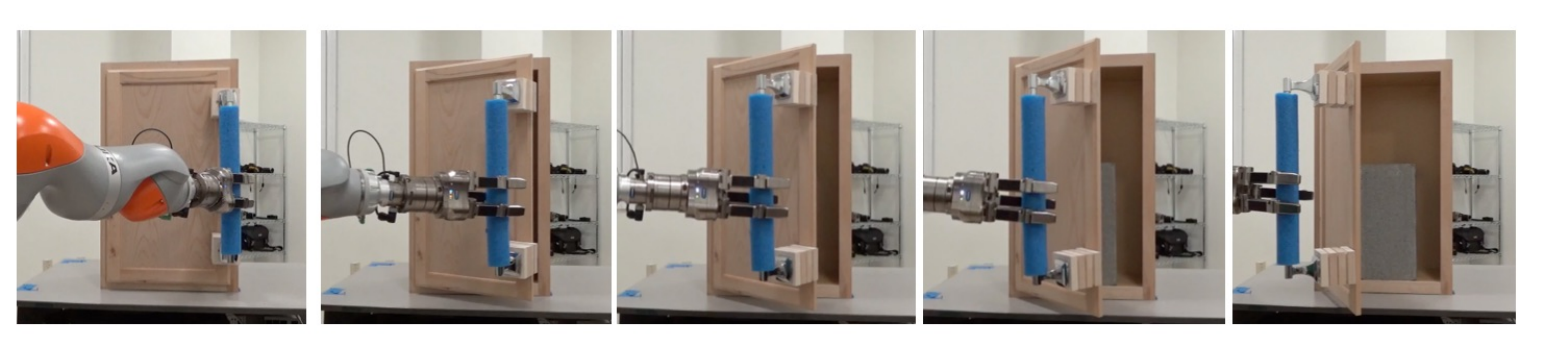}\label{fig:lab_cabinet_open_demo}}

\subfloat[\textit{Turn} Applied to a Novel Box]{\includegraphics[width=1.6\columnwidth]{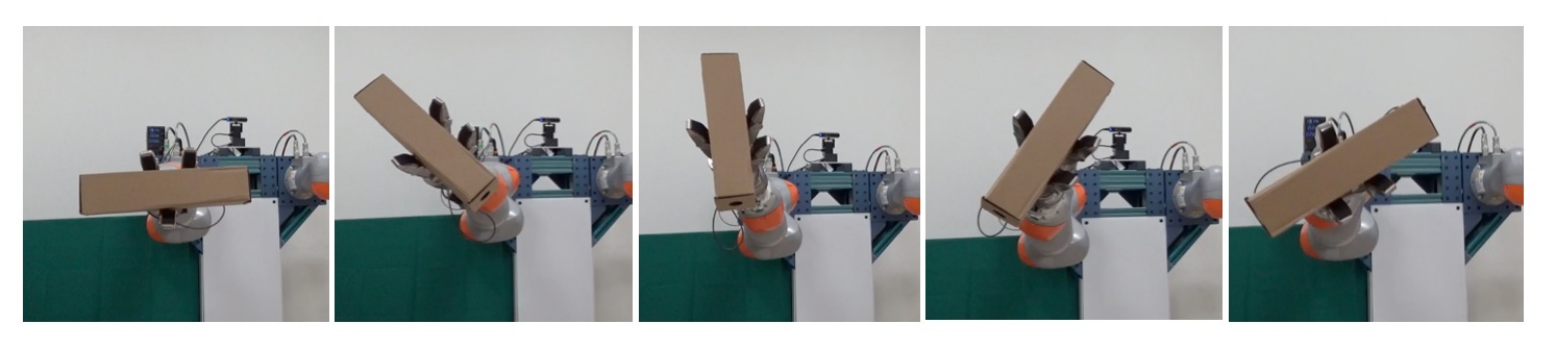}\label{fig:box_turn}}

\subfloat[\textit{TranslateRight} Applied to a Novel Box (only 3 out of 5 generated timesteps shown)]{\includegraphics[width=1.6\columnwidth]{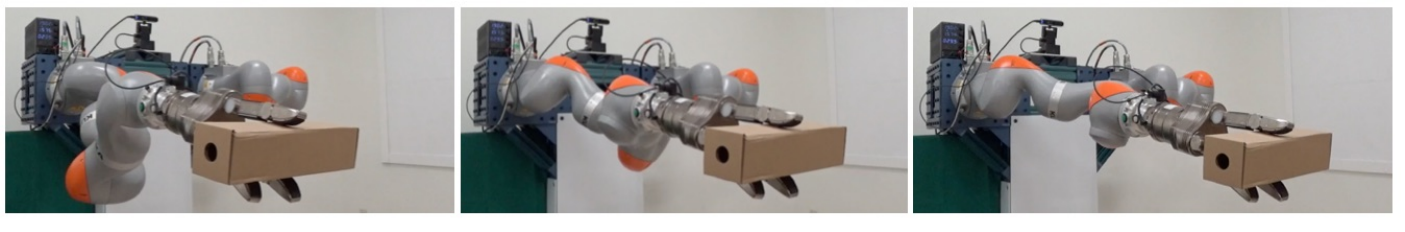}\label{fig:box_right}}

\subfloat[\textit{TranslateLeft} Applied to a Novel Box (only 3 out of 5 generated timesteps shown)]{\includegraphics[width=1.6\columnwidth]{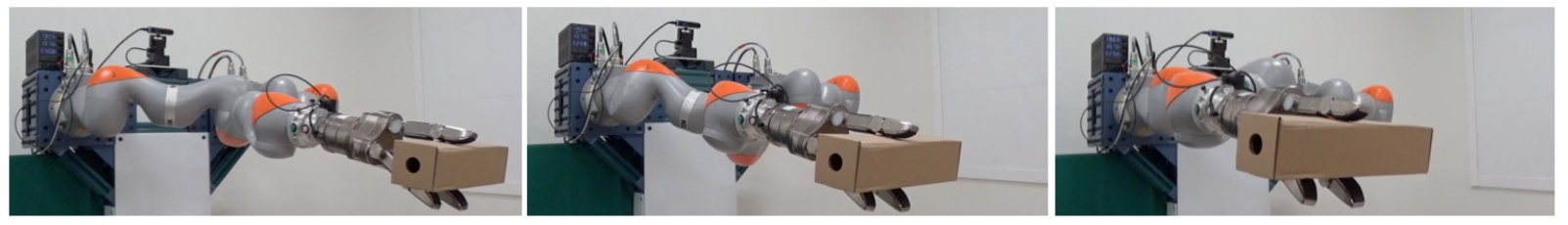}\label{fig:box_left}}

\caption{
Snapshots of robot demonstrations using our model to generate trajectories for motion planning. With our model, the robot is able to manipulate novel object instances (viz., a Box and a Cabinet) for a variety of verbs (\textit{Open}, \textit{Turn}, \textit{TranslateLeft}, and \textit{TranslateRight}).
}
\label{fig:robot-demo}
\end{figure*}

\subsection{Object Trajectory Optimization}

With the trained models from our classifier, we run the CMA-ES optimizer on object instances and qualitatively assess the performance. We set the initial covariance parameter to 0.33, the population size to 40, and the number of generations to 60. Since we trained our CNN classifier with five timesteps sampled at regular intervals (such that the change is equal between chosen timesteps), the optimizer predicts a single change in state (of size $6+n$) that is applied at each timestep to generate the predicted trajectory. Further, we identified that the verbs studied include change in at most one degree of freedom; therefore, to prevent extraneous motion, we take only the maximum predicted per-timestep change in state and set the other dimensions to zero when scoring trajectories. This property is straightforward to compute from the data but could be learned for each verb. The resulting trajectory is rendered and scored as described in Section \ref{sec:trajopt}.  Qualitative results are shown in Figure \ref{fig:sim-demo}.

\subsection{Robot Demonstration}
Finally, we demonstrate that our model enables a real robotic system to execute verb commands on novel object categories.  The robot is a KUKA LBR iiwa7 with a Schunk Dexterous Hand 3-fingered gripper. Five commands were executed: \textit{Open} (applied to a novel instance of the StorageFurniture object category) as well as \textit{Turn}, \textit{TranslateRight}, \textit{TranslateLeft}, and \textit{Push} (applied to a novel instance of the Box object category). After performing the offline trajectory optimization, the robot executes the desired object trajectories using motion planning. Plans for the \textit{Turn}, \textit{Push}, and \textit{Translate} verbs were computed as simple motions in Cartesian space with the object placed in the robot's gripper. The \textit{Open} command was executed by providing the robot with a grasp on the cabinet door (e.g., as though from an off-the-shelf grasp detection algorithm \cite{ten2017grasp}), and computing a trajectory that moves the end-effector to execute the object trajectory produced by our system with MoveIt! \cite{chitta2012moveit}. 

We highlight some of the simulated trajectories that were generated by our method as Figure~\ref{fig:sim-demo}. We also show images of the KUKA robot executing some of the generated trajectories in Figure~\ref{fig:robot-demo}. We provide demonstration videos in our supplementary materials.

\section{Discussion}
\label{sec: Discussion}
In our experiments, we used a small set of manipulation verbs that can be realized either through object trajectory or the difference between initiation and termination states. However, our model can easily incorporate verbs realized through both robot arm trajectory and object arm trajectory by generating RGB trajectory sequences that include both the robot arm and the object (e.g., the verbs \textit{Throw} and \textit{Toss}).

Our implementation allows for flexibility in the length of the object trajectories and the frequency at which they are rendered. For most of our experiments, we used only five out of the total 21 generated timesteps (initiation step, Step 5, Step 10, Step 15, termination step), and we assume that there is a constant amount of time for all the object instances used for training our networks. One could prefer to generate trajectories of length two to only produce the goal/termination state of the verb, but this was observed to decrease classifier performance empirically. One could also adjust the model to consider more timesteps, or pick irregular timesteps (not separated by a constant amount) for each verb. In other words, the number of timesteps can be thought of as way-points for the generated trajectory. Furthermore, we do not consider adverbs in our model; however, there is the possibility that certain adverbs can be incorporated into the model, such as adverbs that describe the speed of an action (e.g., slowly, quickly, carefully), by generating object trajectories where the goal state of a verb is reached at a comparatively earlier or later timestep. 

Due to the flexibility of our model, it is possible that the correct verb goal state is produced even when given a non-canonical initial state of the object, e.g., a door can be opened from a initial state of being slightly ajar rather than completely closed. This could be achieved by running the trajectory optimizer step given a non-canonical state on an object, or training the classifier on more trajectories that incorporate non-canonical initial states.

The training trajectories contain ideal scenarios such as having objects in isolation and single fixed-view angles of each trajectory. To improve our model's performance, more trajectory images taken from different angles may be added to the training data. Furthermore, other state-of-the-art activity and object recognition vision classification methods can be used on trajectory videos or image sequences such as RAANet~\cite{lu2022rangeaware}, YOLO~\cite{shinde2018yolo}, and other methods as described in the survey by \citet{zaidi2022survey}.

\section{Conclusion}

We have proposed a two-part model consisting of a classifier and an optimizer to generalize manipulation skills to novel object categories using verbs. We present a classifier that can recognize which verb is being performed in a given trajectory, and enables verb generalization to new object instances and new object categories. This classifier achieves an average of 76.69\% accuracy over 13 object categories and 14 verbs. The optimizer is responsible for finding kinematic trajectories of an object that scores highly on the classifier for the desired verb command. Our model can generalize skills across novel objects, and we conducted robot demonstrations to show that robots can use our model with motion planning for execution on novel objects.

\section{Acknowledgements}
Special thanks to our mentors and advisors for their support and advising throughout this project. Part of this research was conducted using computational resources and services at the Center for Computation and Visualization, Brown University. This work is supported by NSF under grant numbers IIS-1955361, IIS-1956221, and IIS-1941808; ONR under grant numbers N00014-21-1-2584 and N00014-22-1-2592; AFOSR DURIP FA9550-21-1-0308, and Echo Labs. Disclosure: George Konidaris is the Chief Roboticist of Realtime
Robotics, a robotics company commercializing real-time motion planning.

\bibliographystyle{IEEEtrann}
\bibliography{ref}

\end{document}